\documentclass{article}

\usepackage{amsmath,graphicx,mlspconf}
\usepackage{booktabs}
\usepackage{array}
\usepackage{tabularx}
\usepackage{algorithm}
\usepackage{algorithmic}
\usepackage[caption=false,font=footnotesize]{subfig}
\usepackage{placeins}
\usepackage[table]{xcolor}
\usepackage{multirow}
\usepackage{url}
\usepackage{flushend}
\toappear{2026 IEEE International Workshop on Machine Learning for Signal Processing, Sep.\ 28-- Oct.\ 1, 2026, Atlanta, USA}

\title{Personalized Federated Learning through Global Knowledge Distillation and Local Head Adaptation}

\usepackage[hidelinks]{hyperref}
\usepackage{xcolor}
\usepackage{hyperref}

\name{
Polycarpo Souza Neto$^{1}$,
José Mairton Barros da Silva Jr.$^{2}$,
Charles Casimiro Cavalcante$^{1}$
\thanks{© 2026 IEEE. Personal use of this material is permitted. Permission from IEEE must be obtained for all other uses, in any current or future media, including reprinting/republishing this material for advertising or promotional purposes, creating new collective works, for resale or redistribution to servers or lists, or reuse of any copyrighted component of this work in other works.}
\thanks{This work was partially supported by CNPq Procs. 308512/2023-5
and 381429/2026-1, CNPq/INCT STREAM (Signal Processing and TRansmission
for Environmental Analysis and Monitoring) 409179/2024-8, and by
CAPES -- Finance Code 001.}
\thanks{The source code used in this work has been made available in the
\href{https://github.com/pneto29/pFedKDH_mlsp_26}
{\textcolor{blue}{\underline{source-code repository (github)}}}.}
}

\address{
$^{1}$Universidade Federal do Ceará, Fortaleza, CE, Brazil\\
$^{2}$Uppsala University, Uppsala, Sweden
}

\copyrightnotice{979-8-3195-0884-3/26/\$31.00 {\copyright}2026 IEEE}

\begin{document}
\maketitle

\ninept 

% \begin{abstract}
% Statistical heterogeneity remains a major challenge in personalized federated learning (PFL), often degrading the performance of shared global models. To address this issue, we propose pFedKDH, a personalized federated learning method that combines backbone-only aggregation, persistent client-specific heads, and knowledge distillation from a calibrated global teacher. The method preserves local decision boundaries while transferring global predictive information during local optimization. Experiments on MNIST, FMNIST, CIFAR10, and CIFAR100 under class-wise Dirichlet non-IID partitions show competitive performance against several federated learning baselines, reaching up to 99.28\% accuracy on MNIST, 97.70\% on FMNIST, 92.33\% on CIFAR10, and 68.45\% on CIFAR100. Additional analyses demonstrate the effectiveness of persistent local heads and teacher-guided optimization under heterogeneous data distributions.
% \end{abstract}
\begin{abstract}
Statistical heterogeneity limits federated learning when a single global classifier cannot represent client-specific label distributions. In this work, we propose Personalized Federated Knowledge Distillation with Head Adaptation (pFedKDH), which aggregates only the shared backbone, keeps persistent client-specific heads, and uses a recalibrated global head as a teacher during local training. Across MNIST, Fashion-MNIST, CIFAR10, and CIFAR100 under class-wise Dirichlet partitions, pFedKDH obtains the best accuracy in most settings, with accuracy gaps up to 37.67\% over the weakest baseline and consistently low standard deviation across repetitions. Component-wise diagnostics and convergence results support the role of persistent heads and distillation-guided local optimization under label-skewed data. 
\end{abstract}

\begin{keywords}
Personalized federated learning, knowledge distillation, non-IID data, model personalization.
\end{keywords}

\section{Introduction}
\label{sec_intro}

Federated learning (FL) enables collaborative model training without moving raw client data to a central server~\cite{mcmahan2017communication,kairouz2021advances}. In its standard form, selected clients receive a global model, update it using local data, and send the updated parameters back to the server for aggregation. This setting is attractive for distributed visual learning and edge intelligence, where data are decentralized and may be privacy-sensitive. However, when clients follow different local distributions, server aggregation may mix incompatible class evidence and local classification rules.

% Recent surveys reinforce that FL has been increasingly applied in privacy-sensitive and distributed domains such as healthcare, finance, IoT, edge computing, and decentralized intelligent systems, while non-IID data remains one of the main challenges affecting convergence, communication efficiency, and model robustness \cite{lu2024federated,chung2026decentralized}.

% This limitation becomes critical under statistical heterogeneity, a condition commonly described in federated learning as non-IID data. In this work, label skew is treated as a specific form of non-IID heterogeneity, in which clients differ in their class distributions. More generally, practical FL systems may involve differences in sample size, acquisition conditions, available classes, and label distribution~\cite{ye2023heterogeneous}. Under label skew, local classifiers may learn decision boundaries that are useful for their own data but incompatible with those learned by other clients. Consequently, aggregating the entire model may degrade both global generalization and client-level personalization.
Recent surveys show that FL has been widely explored in privacy-sensitive and distributed domains, including healthcare, finance, IoT, edge computing, and decentralized intelligent systems, but non-IID data remains a central challenge for convergence, communication efficiency, and robustness \cite{lu2024federated,chung2026decentralized}. In this work, we focus on label skew, a common form of statistical heterogeneity in which clients differ in their class distributions~\cite{ye2023heterogeneous}. Under this setting, local classifiers may learn client-specific decision boundaries that become incompatible when the entire model is averaged.

% This limitation becomes critical under statistical heterogeneity, also commonly referred to as non-IID data in federated learning. Here, label skew is considered a specific form of non-IID heterogeneity, where clients exhibit different class distributions. In practical FL systems, clients may differ in sample size, acquisition conditions, available classes, and label distribution~\cite{ye2023heterogeneous}. Under label skew, local classifiers may learn decision boundaries that are useful for their own data but incompatible with those learned by other clients. Consequently, aggregating the entire model may degrade both global generalization and client-level personalization.

Personalized federated learning (PFL) addresses this issue by allowing each client to use a model that differs from a single shared model~\cite{tan2022towards,sabah2023model}. Existing methods regularize personalized models toward a global reference, separate shared and local parameters, adapt aggregation weights, group similar clients, or transfer auxiliary information during training. Although these strategies improve over standard FL, important gaps remain: full-model regularization still couples all parameters to a global reference, adaptive aggregation still mixes model parameters, and backbone-head separation does not necessarily provide a global teacher signal during local optimization.

Our proposed method follows from a simple observation. In visual classification, the backbone learns representations that can be useful across clients, while the classifier head maps these representations to the local label distribution. Therefore, the backbone should be shared, but the classifier head should remain local under strong label skew. At the same time, keeping heads local should not discard global knowledge.

Based on this observation, we introduce Personalized Federated Knowledge Distillation with Head Adaptation (pFedKDH). The method aggregates only the backbone, keeps a persistent client-specific head at each client, and uses a recalibrated global head as a teacher during local training. Global knowledge is transferred through the teacher signal and the shared backbone, rather than through direct aggregation of classifier parameters. The local objective combines supervised learning, distillation, and proximal regularization. After local training, only the updated backbone is returned to the server, while the client-specific head remains persistent across communication rounds.

We evaluate pFedKDH on MNIST, Fashion-MNIST, CIFAR10, and CIFAR100 using class-wise Dirichlet partitions with $\alpha \!\in\! \{0.05, 0.10, 0.50\}$. The comparison includes classical FL baselines and representative personalized methods under a standardized CNN backbone whenever applicable. Across the evaluated settings, pFedKDH achieves the strongest overall accuracy-stability trade-off. It obtains the best mean accuracy in most cases and remains close to the best method in the remaining ones. Moreover, its standard deviation is consistently small compared with competing methods, indicating that the observed gains are stable across repetitions rather than caused by isolated runs. In the settings where pFedKDH leads, the accuracy gap reaches up to 37.67 \% over the weakest compared method and up to 1.82 \% over the strongest competitor. This performance is obtained with a competitive computational cost,  pFedKDH requires a per-round time comparable to FedALA and FedBABU, and substantially lower than FedRep and pFedMe.

The main contributions of this work are summarized as follows:
\begin{itemize}
    \item We introduce pFedKDH, a personalized FL method that avoids classifier-head averaging. The server aggregates only the backbone, while each client keeps a persistent client-specific head that is not uploaded, averaged, or overwritten;

    \item We combine persistent client-specific heads with teacher-guided local optimization. A recalibrated global head acts as a teacher during local training, transferring global knowledge without forcing the client classifier to become global;

    \item We empirically show that this design achieves high accuracy, low variability across repetitions, stable convergence, and competitive computational cost under heterogeneous data partitions.
\end{itemize}

% The rest of the paper is organized as follows. Section~\ref{Sec:Related_works} brings the description of the state-of-art in personalized FL methods and locates our proposal within the literature. The proposal of our method and discussion about its algorithmic implementation is discussed in Section~\ref{sec:method} while the simulation results are presented in Section~\ref{sec:results}. Finally, our conclusions and future investigation steps are drawn in Section~\ref{sec:conclusion}.

\section{Related Work}\label{Sec:Related_works}

Federated learning commonly combines local optimization with server
aggregation to learn a shared model from decentralized data.
FedAvg~\cite{mcmahan2017communication} averages client parameters
after local training, while FedProx~\cite{li2020federated} adds a
proximal term to limit deviations from the global model. Both methods
still rely on a single shared solution, which can be restrictive under
heterogeneous label distributions.

One direction for personalization is to maintain client-specific models while keeping them tied to a global reference. pFedMe~\cite{dinh2020personalized} uses a bilevel formulation in which each client optimizes a personalized model regularized toward the global one. Ditto~\cite{li2021ditto} instead trains global and personalized models in parallel, with a penalty that keeps the local solution close to the global model.
 These methods personalize the full model, so their behavior depends on how strongly the local model is coupled to the global reference. 

A second direction separates the network into a feature extractor and a classifier head. The feature extractor maps the input into a representation, while the head maps this representation to output classes. FedPer~\cite{arivazhagan2019federated} keeps the final layers local and aggregates the lower layers. FedRep~\cite{collins2021exploiting} separates shared representation learning from local classifier training. FedBABU~\cite{oh2022fedbabu} updates and aggregates only the body of the network during federated training and adapts the classifier afterward. These methods show that classifier parameters are strongly affected by local label distributions and should not necessarily be averaged together with the shared representation. Fed-RoD~\cite{chen2022fedrod} combines generic
and personalized predictors, while FedLoGe \cite{xiao2024fedloge}
 uses shared repre-
sentations with client-specific classifiers under long-tailed data.

Another group of methods modifies the information exchanged across clients or the way client updates are combined. FedProto~\cite{tan2022fedproto} exchanges one feature summary per class instead of full model parameters, reducing the dependence on full-model averaging. This mechanism can be weakened when local data are sparse or when some classes are absent from a client. pFedSim~\cite{tan2023pfedsim} aggregates information from clients estimated to be similar, which reduces the influence of unrelated updates but depends on the reliability of the similarity estimate. FedALA~\cite{zhang2023fedala} learns how much of the downloaded global model should be combined with the previous local model before local training, giving each client more control over the use of global information. However, these approaches still depend on class summaries, similarity estimates, or parameter mixing.

Knowledge distillation offers another way to transfer information across
heterogeneous clients. FedFed~\cite{yang2023fedfed} uses feature
distillation to mitigate data heterogeneity while keeping part of the
information local. pFedKDH follows this direction while keeping shared representation
learning separate from local classification. Only the backbone is
aggregated, whereas each client keeps a private and persistent head.
Global predictive information is transferred through a recalibrated
teacher, preserving client-specific decision boundaries without head
aggregation or replacement.

\section{Proposed Method}
\label{sec:method}

\begin{figure}[h]
\newcommand{\archheight}{0.35\textheight}

\centering
\includegraphics[
    width=0.92\columnwidth,
    height=\archheight,
    keepaspectratio
]{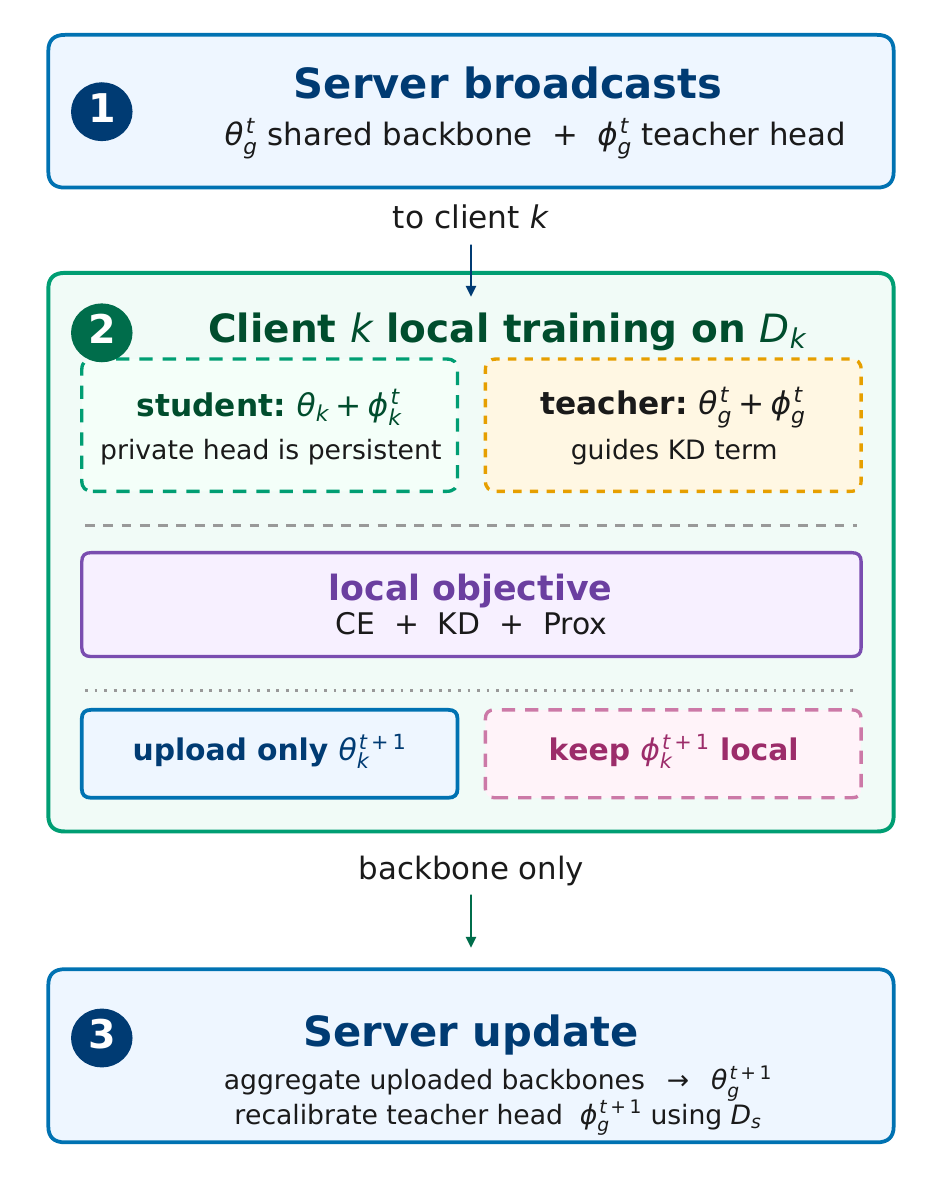}
\caption{Overview of the pFedKDH training cycle: the server broadcasts the shared backbone and teacher head, while each client performs local training with CE, KD, and proximal regularization. Only the updated backbone is uploaded to the server; private heads remain local, and the teacher head is recalibrated using server-side data.}
\label{archt}
\end{figure}

This section presents our method. The pFedKDH separates shared representation learning from client-specific classification under label skew. It aggregates a global backbone while keeping persistent private heads, and uses a calibrated global teacher for local distillation. A proximal term limits backbone drift without exposing client-specific classifiers. Figure~\ref{archt} summarizes the proposed pFedKDH workflow.

\subsection{Problem Setup and Model Decomposition}

We consider a FL setting with $K$ clients, where each client $k$ owns a private dataset $\mathcal{D}_k=\{(x_i^k,y_i^k)\}_{i=1}^{n_k}$ and $n_k$ denotes the number of local training samples. Each client also maintains a private and persistent local head $\phi_k$,
while the server maintains the global backbone $\theta_g$, the teacher
head $\phi_g$, and the auxiliary labeled dataset $\mathcal{D}_s$.
The model is decomposed as \(f(x;\theta,\phi)=h_{\phi}(b_{\theta}(x))\), where \(\theta\) denotes the backbone parameters, \(\phi\) denotes the classification-head parameters, \(b_{\theta}(\cdot)\) maps the input to a feature representation, and \(h_{\phi}(\cdot)\) maps this representation to class logits.

pFedKDH can be formulated as a personalized federated optimization problem in which a shared representation \(\theta_g\) is learned together with client-specific heads \(\{\phi_k\}_{k=1}^{K}\). The corresponding global objective aggregates the local losses weighted by the relative number of samples at each client:
\begin{equation}
\label{eq:pfedkdh_personalized_objective}
    \min_{\theta_g,\{\phi_k\}_{k=1}^{K}}
    \sum\nolimits_{k=1}^{K}
    \frac{n_k}{\sum_{\ell=1}^{K}n_{\ell}}
    \mathcal{L}_k(\theta_g,\phi_k).
\end{equation}
In practice, this objective is optimized in a federated manner: each client updates only its local copy of the backbone and its private head, while the server aggregates the backbone updates. This formulation encodes the main design choice of pFedKDH: the backbone captures transferable visual structure, while the local heads absorb client-specific label distributions.

At round $t$, the server maintains a global model $w_g^t=(\theta_g^t,\phi_g^t)$, where $\theta_g^t$ is the global backbone and $\phi_g^t$ is the global head used by the teacher. Client $k$ maintains a personalized model $w_k^t=(\theta_g^t,\phi_k^t)$, where $\phi_k^t$ is private and persistent across rounds. The server coordinates representation learning through \(\theta_g^t\), while each client preserves and updates its own persistent head \(\phi_k^t\) across rounds.

\subsection{Local Personalized Training}

At round $t$, the server selects clients $\mathcal{S}_t$ according to the participation ratio $\rho$ and broadcasts the current global model $(\theta_g^t,\phi_g^t)$. Each selected client initializes its local backbone as $\theta_k \leftarrow \theta_g^t$ and its classifier as the persistent local head $\phi_k^t$.

The global model acts as a teacher, while the personalized client
model acts as a student. Following knowledge distillation
\cite{hinton2015distilling}, the predictive distribution associated
with logits $\mathbf{z}$ is softened using temperature $T$ as
$q_i^T=\exp(z_i/T)/\sum_j \exp(z_j/T)$, where $T>1$ produces a
smoother distribution that exposes relative information among
classes. Let $q_g^T(x)$ and $q_k^T(x)$ denote the resulting teacher
and student distributions, respectively. Client $k$ minimizes
% \begin{equation}
% \label{eq:pfedkdh_local_objective}
% \begin{split}
%     \mathcal{J}_k^t(\theta_k,\phi_k)
%     =
%     &
%     \frac{1-\lambda_{\mathrm{KD}}}{n_k}
%     \sum_{(x,y)\in\mathcal{D}_k}
%     \ell_{\mathrm{CE}}
%     \left(h_{\phi_k}(b_{\theta_k}(x)),y\right)
%     \\
%     &
%     +
%     \frac{\lambda_{\mathrm{KD}}T^2}{n_k}
%     \sum_{(x,y)\in\mathcal{D}_k}
%     \mathrm{KL}
%     \left(q_g^T(x)\,\|\,q_k^T(x)\right)
%     +
%     \frac{\mu}{2}
%     \left\|\theta_k-\theta_g^t\right\|_2^2 .
% \end{split}
% \end{equation}

\begin{equation}
\label{eq:pfedkdh_local_objective}
\begin{split}
    \mathcal{J}_k^t(\theta_k,\phi_k)
    =
    &
    \frac{1-\lambda_{\mathrm{KD}}}{n_k}
    \sum_{(x,y)\in\mathcal{D}_k}
    \ell_{\mathrm{CE}}
    \left(h_{\phi_k}(b_{\theta_k}(x)),y\right)
    \\
    &
    +
    \frac{\lambda_{\mathrm{KD}}T^2}{n_k}
    \sum_{(x,y)\in\mathcal{D}_k}
    \mathrm{KL}
    \left(q_g^T(x)\,\|\,q_k^T(x)\right)
    \\
    &
    +
    \frac{\mu}{2}
    \left\|\theta_k-\theta_g^t\right\|_2^2 .
\end{split}
\end{equation}

The three terms fit local labels, transfer global predictive knowledge,
and constrain backbone drift under non-IID updates. Distillation guides
the local model without uploading, averaging, or overwriting the
client-specific head, preserving the personalized decision boundary.
After local training, only $\theta_k^{t+1}$ is sent to the server, while
$\phi_k^{t+1}$ remains private and persistent.

\subsection{Server-Side Warm-up, Aggregation, and Teacher Recalibration}

pFedKDH uses two server-side operations: global warm-up and head recalibration. We define global warm-up as the optional pre-training of the global model \(w_g^0=(\theta_g^0,\phi_g^0)\) before the first communication round. In this step, the server updates both the global backbone \(\theta_g^0\) and the global head \(\phi_g^0\) using the auxiliary server dataset \(\mathcal{D}_s\):
\begin{equation}
\label{eq:pfedkdh_warmup}
    (\theta_g^0,\phi_g^0)
    \leftarrow
    \arg\min_{\theta_g,\phi_g}
    \frac{1}{|\mathcal{D}_s|}
    \sum_{(x,y)\in\mathcal{D}_s}
    \ell_{\mathrm{CE}}
    \left(
        h_{\phi_g}(b_{\theta_g}(x)),y
    \right).
\end{equation}

This global warm-up is distinct from head recalibration: warm-up updates both \(\theta_g^0\) and \(\phi_g^0\) before federated training starts, whereas head recalibration updates only the global head while keeping the current global backbone fixed.

% This stage prevents the first distillation steps from relying on an entirely untrained teacher. When $E_w=0$, the warm-up is skipped; this is the setting used in the main fair comparison.

We denote by \(E_w\) the number of global warm-up epochs. When used, the global warm-up initializes the teacher before distillation.

After local training, the server aggregates only the client backbones:
\begin{equation}
\label{eq:pfedkdh_backbone_aggregation}
    \theta_g^{t+1}
    =
    \sum\nolimits_{k\in\mathcal{S}_t}
    \frac{n_k}
    {\sum_{\ell\in\mathcal{S}_t} n_{\ell}}
    \theta_k^{t+1}.
\end{equation}
No client head is transmitted, averaged, or reset. This is the core personalization mechanism of pFedKDH: the representation is collaborative, but the decision boundary remains local.

% Because the aggregated backbone changes after each round, the previous global head may no longer be well aligned with the updated representation. The server therefore recalibrates only the global head, keeping $\theta_g^{t+1}$ fixed:
% \begin{equation}
% \label{eq:pfedkdh_head_recalibration}
%     \phi_g^{t+1}
%     \leftarrow
%     \arg\min_{\phi}
%     \frac{1}{|\mathcal{D}_s|}
%     \sum_{(x,y)\in\mathcal{D}_s}
%     \ell_{\mathrm{CE}}
%     \left(
%         h_{\phi}(b_{\theta_g^{t+1}}(x)),y
%     \right).
% \end{equation}
% Thus, warm-up and recalibration have different purposes. Warm-up initializes the full global model before federated training, whereas recalibration realigns only the teacher head after each backbone aggregation. The auxiliary dataset $\mathcal{D}_s$ is used only for teacher initialization and calibration, and never for testing.

Global warm-up starts before the first communication round, only when \(E_w\!>\!0\). In this stage, the server updates the initial global backbone \(\theta_g^0\) and global head \(\phi_g^0\) on the auxiliary dataset \(\mathcal{D}_s\), before any client receives the global model.

Head recalibration starts after each server-side backbone aggregation step. Because the aggregated backbone \(\theta_g^{t+1}\) changes at the end of round \(t\), the previous global head may no longer be well aligned with the updated representation. The server therefore recalibrates only the global head, keeping \(\theta_g^{t+1}\) fixed:
\begin{equation}
\label{eq:pfedkdh_head_recalibration}
    \phi_g^{t+1}
    \leftarrow
    \arg\min_{\phi}
    \frac{1}{|\mathcal{D}_s|}
    \sum_{(x,y)\in\mathcal{D}_s}
    \ell_{\mathrm{CE}}
    \left(
        h_{\phi}(b_{\theta_g^{t+1}}(x)),y
    \right).
\end{equation}

Thus, warm-up and recalibration occur at different moments and serve different purposes. Warm-up is an optional pre-federated stage that initializes the full global model before round \(0\), whereas recalibration is a repeated post-aggregation stage that realigns only the teacher head after each communication round. The auxiliary dataset \(\mathcal{D}_s\) is used only for teacher initialization and calibration, and never for testing.

Algorithm~\ref{alg:pfedkdh} summarizes the implementation of our method.

%\subsection{Algorithmic Summary}

% Algorithm~\ref{alg:pfedkdh} summarizes pFedKDH. The server broadcasts the global backbone and global teacher head to selected clients. Each client updates a local backbone and its persistent head using supervised learning, distillation, and proximal regularization. Only the updated backbone returns to the server. The server then performs weighted backbone aggregation and recalibrates the global teacher head.

\begin{algorithm}[!h]
\caption{pFedKDH}
\label{alg:pfedkdh}
\begin{algorithmic}[1]
\REQUIRE Number of clients $K$, rounds $R$, local epochs $E$, participation ratio $\rho$, distillation weight $\lambda_{\mathrm{KD}}$, temperature $T$, proximal weight $\mu$, auxiliary dataset $\mathcal{D}_s$, warm-up epochs $E_w$
\STATE Initialize global backbone $\theta_g^0$, global head $\phi_g^0$, and persistent local heads $\{\phi_k^0\}_{k=1}^{K}$
\IF{$E_w > 0$}
    \STATE Warm up $(\theta_g^0,\phi_g^0)$ on $\mathcal{D}_s$ using \eqref{eq:pfedkdh_warmup}
\ENDIF
\STATE Calibrate $\phi_g^0$ on $\mathcal{D}_s$ with $\theta_g^0$ fixed
\FOR{$t=0,1,\ldots,R-1$}
    \STATE Select clients $\mathcal{S}_t \subseteq \{1,\ldots,K\}$ according to $\rho$
    \STATE Broadcast $(\theta_g^t,\phi_g^t)$ to each client $k\in\mathcal{S}_t$
    \FOR{each client $k \in \mathcal{S}_t$ in parallel}
        \STATE Initialize $\theta_k \leftarrow \theta_g^t$ and $\phi_k \leftarrow \phi_k^t$
        \STATE Update $(\theta_k,\phi_k)$ by minimizing \eqref{eq:pfedkdh_local_objective} for $E$ local epochs
        \STATE Send $\theta_k^{t+1}$ to the server
        \STATE Keep $\phi_k^{t+1}$ private and persistent
    \ENDFOR
    \STATE Aggregate $\theta_g^{t+1}$ using \eqref{eq:pfedkdh_backbone_aggregation}
    \STATE Recalibrate $\phi_g^{t+1}$ on $\mathcal{D}_s$ with $\theta_g^{t+1}$ fixed using \eqref{eq:pfedkdh_head_recalibration}
\ENDFOR
\end{algorithmic}
\end{algorithm}

% \subsection{Personalized Evaluation and Variants}

% At evaluation time, the global model is $(\theta_g^t,\phi_g^t)$, while the personalized model for client $k$ is $(\theta_g^t,\phi_k^t)$. We report two variants. \textbf{pFedKDH-Fair} evaluates the persistent local head directly, without any extra adaptation. This variant isolates the effect of backbone-only aggregation, persistent heads, distillation, and recalibration under the same training budget used by the baselines.

% \textbf{pFedKDH-EA} adds evaluation-time adaptation. Starting from the same global backbone and persistent local head, the method performs a small number of head-only updates on the client's local training data, while keeping the backbone frozen. No test sample is used during this stage. This variant measures how much additional personalization can be obtained through a lightweight local refinement of the classifier head.
\subsection{Personalized Evaluation and Variants}

At evaluation time, the global model is $(\theta_g^t,\phi_g^t)$, while the personalized model for client $k$ is $(\theta_g^t,\phi_k^t)$. We report two variants:

\begin{enumerate}
    \item \textbf{pFedKDH-Fair:} evaluates the persistent local head directly, without any extra local update after training. This variant isolates the effect of backbone-only aggregation, persistent heads, distillation, and recalibration under the same training budget used by the baselines.

    \item \textbf{pFedKDH-EA:} performs evaluation-time head adaptation through a small number of additional local head-only update steps on the client's training data. Starting from the same global backbone and persistent local head, only $\phi_k^t$ is updated, while $\theta_g^t$ remains frozen. No test sample is used during this stage. This variant measures how much additional personalization can be obtained through a lightweight refinement of the classifier head.
\end{enumerate}
\vspace{-1.5mm}

\section{Experiments and Results}
\label{sec:results}

\subsection{Experimental Protocol}
\label{subsec:experimental_protocol}

All methods were evaluated on four image classification datasets, namely, MNIST, Fashion-MNIST (FMNIST), CIFAR10, and CIFAR100. These benchmarks represent increasingly difficult visual classification scenarios, ranging from grayscale digit recognition to natural image classification with a large number of classes.

Statistical heterogeneity was simulated using class-wise Dirichlet partitions with $\alpha \in \{0.05,0.10,0.50\}$, where smaller values indicate stronger label skew across clients.

The comparison includes classical federated baselines and representative personalized FL methods. Specifically, we compare against FedAvg, FedProx, Ditto, pFedMe, FedPer, FedRep, FedBABU, FedProto, FedALA, FedFed, and pFedSim.

To ensure a fair comparison, all methods were evaluated with the same lightweight CNN architecture whenever applicable. The backbone contains two convolutional layers with ReLU activations and max-pooling, followed by dropout, flattening, and a fully connected projection layer that produces the feature representation used by a linear classification head. Therefore, the observed differences are primarily attributed to the federated optimization and personalization strategies rather than to variations in model capacity.

All methods were trained under the same optimization budget whenever applicable, including the same number of communication rounds, client participation ratio, local epochs, batch size, and learning rate. The main experimental settings are summarized in Table~\ref{tab:experimental_setup}. In all main experiments reported for both pFedKDH-Fair and
pFedKDH-EA, we set $E_w=0$; therefore, the global warm-up stage
in (3) is disabled and $D_s$ is not used to pretrain the global
backbone.

% For pFedKDH, each client keeps a persistent client-specific head across communication rounds. This head is trained locally, but it is not uploaded to the server, not averaged with other heads, and not overwritten by the global model. The server aggregates only the backbone parameters. In addition, pFedKDH uses an auxiliary calibration set $D_s$ with 5000 samples to recalibrate the global head after backbone aggregation. The auxiliary set \(D_s\) is sampled from the training split and kept disjoint from all client test sets. It is used only for teacher initialization and head recalibration, and it is never used to update personalized heads or to report test accuracy. This recalibrated global head is used only as a teacher during local training and is never used as test data. The optional global warm-up stage is disabled in the main experiments to avoid giving pFedKDH an additional initialization advantage. Warm-up is used only in the dedicated comparison reported separately.

% The evaluation protocol follows the nature of each method: shared-model baselines are evaluated as global models, while personalized methods are evaluated at the client level. The detailed metric definitions and result analysis are presented in Section~\ref{subsec:accuracy}.

For the experimental comparison, pFedKDH follows the training procedure defined in Section~\ref{sec:method}. Shared-model baselines are evaluated as global models, whereas personalized methods are evaluated at the client level using their personalized models. For pFedKDH, this corresponds to evaluating the global backbone with each client's persistent head. The detailed metric definitions and result analysis are presented in Section~\ref{subsec:accuracy}.

\begin{table}[!t]
\centering
\caption{Experimental setup used for pFedKDH and baseline comparisons.}
\label{tab:experimental_setup}
\scriptsize
\setlength{\tabcolsep}{4pt}
\renewcommand{\arraystretch}{1.10}
\begin{tabularx}{\columnwidth}{
>{\raggedright\arraybackslash}p{0.42\columnwidth}
>{\raggedright\arraybackslash}X
}
\hline
\textbf{Parameter} & \textbf{Value} \\
\hline
Datasets & MNIST, Fashion-MNIST, CIFAR10, CIFAR100 \\
Data partitioning & Class-wise Dirichlet \\
Dirichlet \(\alpha\) & \(\{0.05, 0.10, 0.50\}\) \\
Number of clients & 20 \\
Client participation & 50\% \\
Communication rounds & 50, 50, 50, 70 \\
Local epochs & 2, 2, 3, 5 \\
Batch size & 64 \\
Learning rate & 0.015 \\
Model architecture & Simple CNN + linear head \\
Aggregated parameters & Backbone only \\
Client-specific heads & Persistent \\
Auxiliary set size \(|\mathcal{D}_s|\) & 5000 \\
Global warm-up & \(E_w=0\) \\
Head recalibration & 3 epochs \\
\hline
\end{tabularx}
\end{table}

\vspace{-1.mm}

\subsection{Convergence Behavior}
\label{subsec:convergence_behavior}

% \begin{figure*}%[!h]
% \centering
% \subfloat[CIFAR10, $\alpha=0.05$]{
%     \includegraphics[width=0.45\linewidth]{figures_mlsp/cifar10_alpha_0p05_mean_std_band_compact.png}
% }
% %\vspace{0.10cm}
% \hfill
% \subfloat[FMNIST, $\alpha=0.05$]{
%     \includegraphics[width=0.45\linewidth]{figures_mlsp/fmnist_alpha_0p05_mean_std_band_compact.png}
% }

% \caption{Accuracy curves over communication rounds for CIFAR10 and FMNIST under the strongest statistical heterogeneity setting considered in this work ($\alpha=0.05$). Shaded regions indicate the standard deviation across repetitions.}
% \label{fig:convergence_alpha005}
% \end{figure*}

\begin{figure*}[!t]
\centering

\subfloat[CIFAR10, $\alpha=0.05$]{%
    \includegraphics[width=0.45\textwidth]{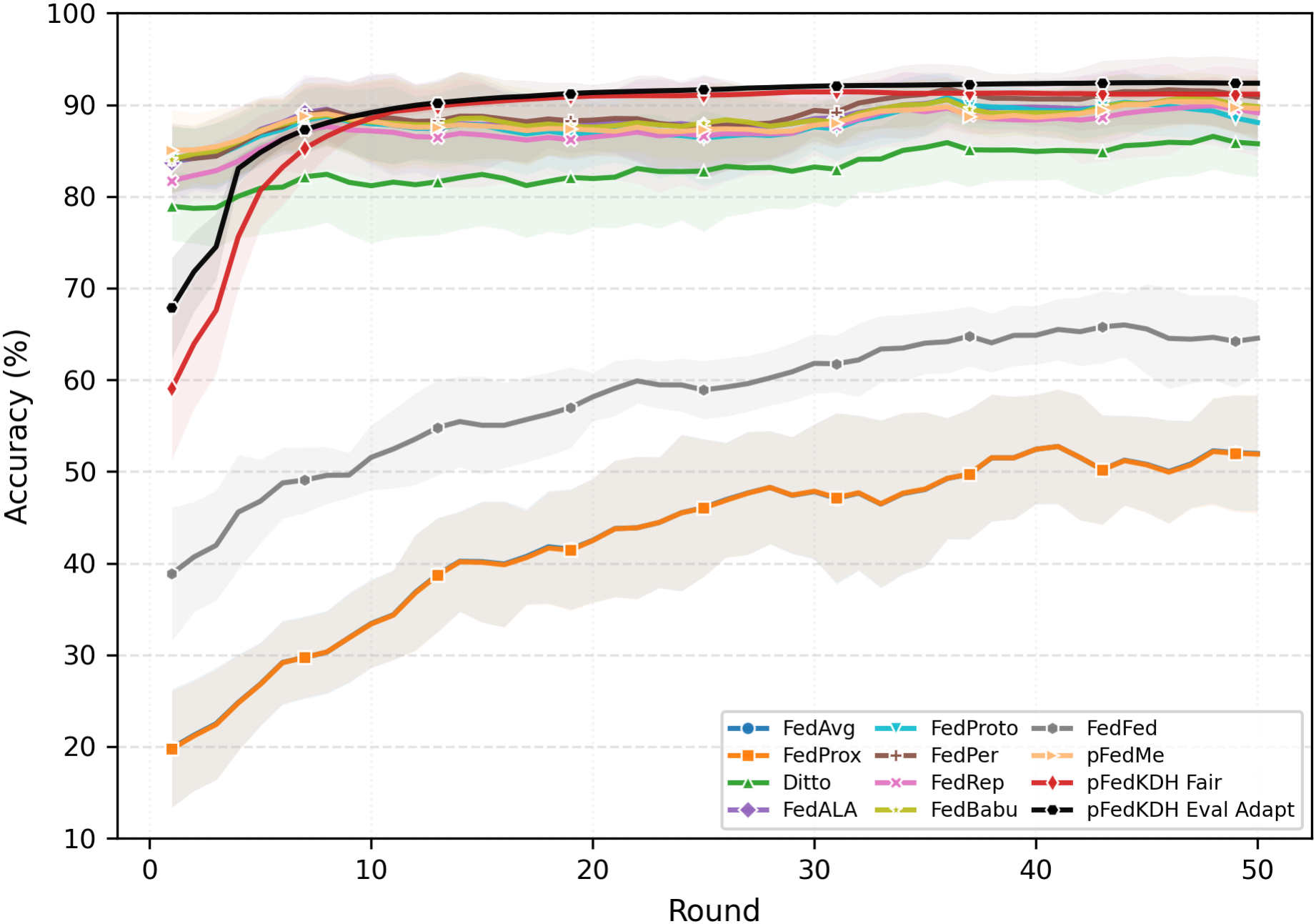}
}
\hfill
\subfloat[FMNIST, $\alpha=0.05$]{%
    \includegraphics[width=0.45\textwidth]{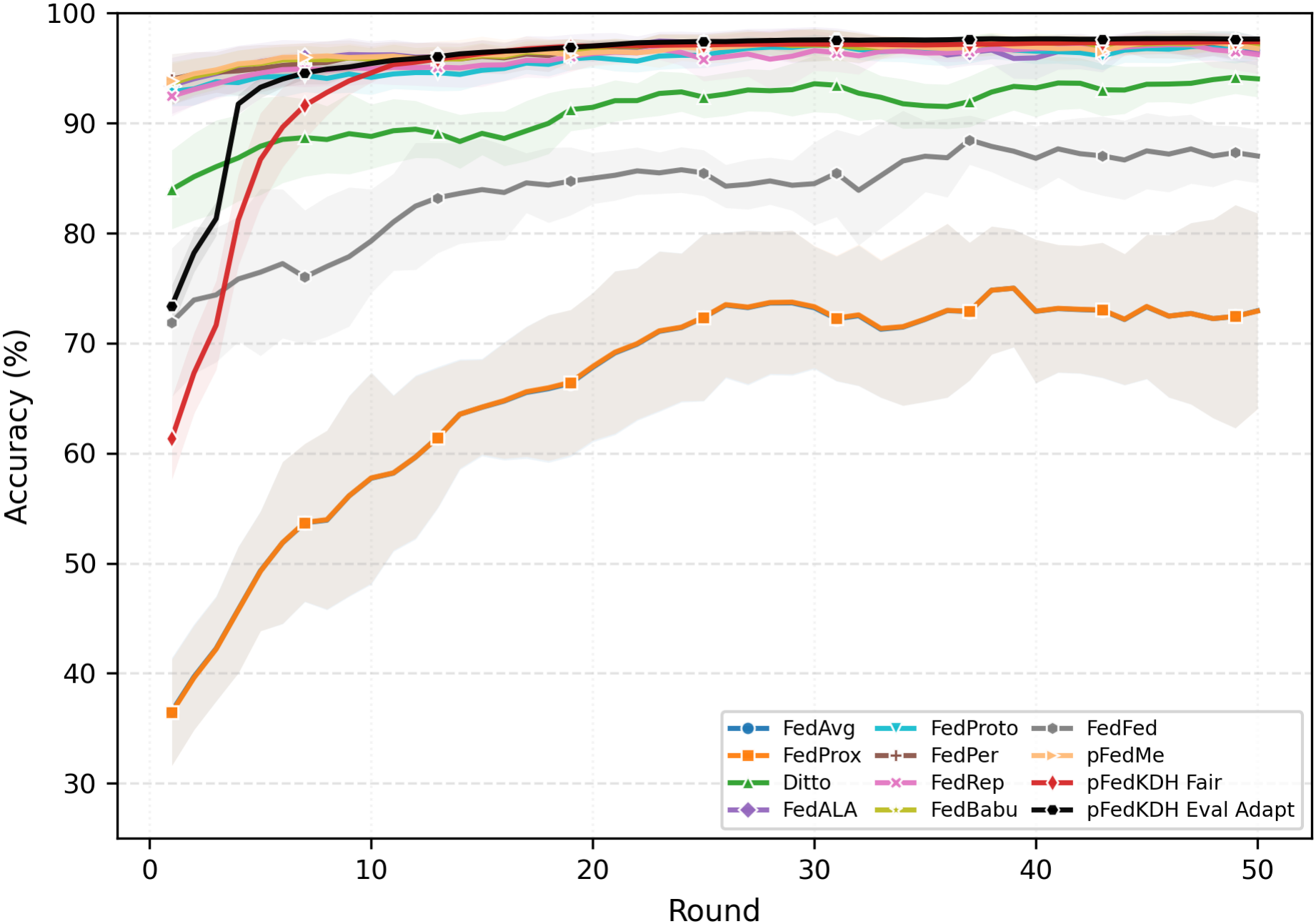}
}

\caption{Accuracy curves over communication rounds for CIFAR10 and FMNIST under the strongest statistical heterogeneity setting considered in this work ($\alpha=0.05$). Shaded regions indicate the standard deviation across repetitions.}
\label{fig:convergence_alpha005}
\end{figure*}

Figure~\ref{fig:convergence_alpha005} analyzes the convergence behavior under the most severe label-skew setting considered in this work. 
% These curves complement the final accuracy tables by showing that the performance of pFedKDH is not caused by isolated final-round fluctuations. Instead, both pFedKDH variants reach the top accuracy range during training and maintain stable trajectories across communication rounds.
On CIFAR10, pFedKDH-EA presents the strongest convergence profile, reaching the highest final accuracy while keeping a narrow standard deviation band. The base version, pFedKDH-Fair, also remains competitive throughout training, showing that the core mechanism of backbone-only aggregation with persistent client-specific heads already provides a stable personalized solution. The additional evaluation-time head adaptation further improves the final client-level performance. Regarding Fashion-MNIST, pFedKDH-EA again converges to the best performance region with low variability across repetitions. In contrast, global baselines such as FedAvg and FedProx converge to substantially lower plateaus, which reinforces the limitation of a single shared classifier under strong statistical heterogeneity. Overall, the convergence curves show that pFedKDH combines high final accuracy with stable training behavior, supporting the robustness of persistent heads and teacher-guided local optimization under severe label skew.

\subsection{Accuracy under Statistical Heterogeneity}
\label{subsec:accuracy}

We report two main evaluation metrics. Global test accuracy measures the performance of a single shared model after server aggregation and is used for shared-model baselines such as FedAvg and FedProx. Personalized macro accuracy measures client-level personalization. It first computes the test accuracy of each client using its own personalized model and then averages these accuracies with equal weight across clients. This metric gives the same importance to all clients and is therefore more suitable for assessing personalized methods under heterogeneous label distributions. Values are reported as mean\(\pm\)standard deviation, and the best result in each setting is shown in bold.

\begin{table}[!t]
\centering
\scriptsize
\setlength{\tabcolsep}{2.5pt}
\renewcommand{\arraystretch}{1.06}

% =========================
% TABLE 2
% =========================
\caption{Final accuracy comparison on MNIST and Fashion-MNIST.}
\label{tab:accuracy_mnist_fmnist}
\resizebox{\columnwidth}{!}{%
\begin{tabular}{l|ccc|ccc}
\toprule
\textbf{Method}
& \multicolumn{3}{c|}{\textbf{MNIST}}
& \multicolumn{3}{c}{\textbf{Fashion-MNIST}} \\
& $\alpha=0.05$ & $\alpha=0.10$ & $\alpha=0.50$
& $\alpha=0.05$ & $\alpha=0.10$ & $\alpha=0.50$ \\
\midrule
FedAvg~\cite{mcmahan2017communication}
& 93.23$\pm$2.11 & 94.43$\pm$2.01 & 97.89$\pm$0.03
& 73.21$\pm$9.79 & 80.96$\pm$6.55 & 86.53$\pm$1.57 \\

FedProx~\cite{li2020federated}
& 93.22$\pm$2.08 & 94.46$\pm$1.96 & 97.88$\pm$0.03
& 73.23$\pm$9.77 & 80.99$\pm$6.45 & 86.49$\pm$1.55 \\
\midrule

Ditto~\cite{li2021ditto}
& 97.47$\pm$0.67 & 96.02$\pm$1.45 & 94.04$\pm$1.35
& 94.42$\pm$0.22 & 91.67$\pm$2.73 & 88.25$\pm$0.36 \\

FedALA~\cite{zhang2023fedala}
& 99.18$\pm$0.26 & 99.05$\pm$0.25 & 95.31$\pm$5.03
& 97.01$\pm$0.40 & 94.96$\pm$2.01 & \textbf{93.25$\pm$0.53} \\

FedProto~\cite{tan2022fedproto}
& 99.00$\pm$0.38 & 97.98$\pm$0.75 & 96.38$\pm$0.73
& 96.76$\pm$1.21 & 95.25$\pm$0.90 & 91.47$\pm$0.35 \\

FedPer~\cite{arivazhagan2019federated}
& 99.11$\pm$0.36 & 98.61$\pm$0.13 & 98.08$\pm$0.30
& 97.10$\pm$0.46 & 94.95$\pm$2.13 & 92.55$\pm$0.35 \\

FedRep~\cite{collins2021exploiting}
& 98.63$\pm$0.53 & 96.69$\pm$1.89 & 96.52$\pm$1.59
& 94.55$\pm$3.38 & 94.14$\pm$1.69 & 91.68$\pm$0.17 \\

FedBABU~\cite{oh2022fedbabu}
& 99.10$\pm$0.42 & 98.49$\pm$0.25 & 97.73$\pm$0.13
& 96.78$\pm$0.29 & 95.18$\pm$1.72 & 92.79$\pm$0.25 \\

FedFed~\cite{yang2023fedfed}
& 98.19$\pm$0.28 & 97.59$\pm$0.74 & 97.63$\pm$0.14
& 84.06$\pm$1.96 & 85.55$\pm$2.68 & 87.54$\pm$1.11 \\

pFedMe~\cite{dinh2020personalized}
& 99.20$\pm$0.23 & 98.76$\pm$0.23 & 97.58$\pm$0.60
& 96.77$\pm$0.57 & 94.75$\pm$1.66 & 92.05$\pm$1.01 \\
\midrule

pFedKDH-Fair
& 99.22$\pm$0.17 & 99.12$\pm$0.09 & 98.96$\pm$0.09
& 97.31$\pm$0.11 & 95.96$\pm$0.27 & 92.13$\pm$0.28 \\

pFedKDH-EA
& \textbf{99.28$\pm$0.03} & \textbf{99.23$\pm$0.03} & \textbf{99.01$\pm$0.09}
& \textbf{97.70$\pm$0.20} & \textbf{96.80$\pm$0.09} & 92.68$\pm$0.30 \\
\bottomrule
\end{tabular}%
}

\vspace{2mm}

% =========================
% TABLE 3
% =========================
\caption{Final accuracy comparison on CIFAR10 and CIFAR100.}
\label{tab:accuracy_cifar}
\resizebox{\columnwidth}{!}{%
\begin{tabular}{l|ccc|ccc}
\toprule
\textbf{Method}
& \multicolumn{3}{c|}{\textbf{CIFAR10}}
& \multicolumn{3}{c}{\textbf{CIFAR100}} \\
& $\alpha=0.05$ & $\alpha=0.10$ & $\alpha=0.50$
& $\alpha=0.05$ & $\alpha=0.10$ & $\alpha=0.50$ \\
\midrule
FedAvg~\cite{mcmahan2017communication}
& 54.67$\pm$8.16 & 61.46$\pm$5.94 & 67.52$\pm$1.37
& 35.74$\pm$0.76 & 38.32$\pm$0.87 & 43.76$\pm$0.15 \\

FedProx~\cite{li2020federated}
& 54.66$\pm$8.27 & 61.44$\pm$5.86 & 67.51$\pm$1.36
& 35.64$\pm$0.76 & 38.25$\pm$0.90 & 43.70$\pm$0.05 \\
\midrule

Ditto~\cite{li2021ditto}
& 86.05$\pm$4.07 & 82.32$\pm$3.19 & 68.82$\pm$2.93
& 57.02$\pm$0.54 & 50.90$\pm$3.28 & 40.18$\pm$1.40 \\

FedALA~\cite{zhang2023fedala}
& 88.63$\pm$4.06 & 87.92$\pm$2.39 & 74.69$\pm$4.45
& 67.53$\pm$2.05 & \textbf{60.86$\pm$3.75} & \textbf{49.64$\pm$1.36} \\

FedProto~\cite{tan2022fedproto}
& 88.35$\pm$3.22 & 84.16$\pm$3.53 & 71.02$\pm$1.24
& 63.80$\pm$1.55 & 56.43$\pm$1.16 & 35.98$\pm$0.99 \\

FedPer~\cite{arivazhagan2019federated}
& 90.51$\pm$4.45 & 87.24$\pm$4.03 & 75.46$\pm$2.13
& 67.51$\pm$2.11 & 60.04$\pm$1.60 & 42.36$\pm$0.52 \\

FedRep~\cite{collins2021exploiting}
& 89.21$\pm$5.29 & 86.49$\pm$3.13 & 70.36$\pm$2.64
& 60.72$\pm$1.34 & 54.03$\pm$1.61 & 36.60$\pm$0.55 \\

FedBABU~\cite{oh2022fedbabu}
& 88.61$\pm$3.36 & 87.08$\pm$2.23 & 76.26$\pm$1.05
& 63.36$\pm$1.34 & 57.31$\pm$2.82 & 44.90$\pm$0.63 \\

FedFed~\cite{yang2023fedfed}
& 65.67$\pm$2.34 & 66.04$\pm$3.81 & 62.56$\pm$0.32
& 36.44$\pm$0.49 & 35.10$\pm$1.34 & 34.93$\pm$0.07 \\

pFedMe~\cite{dinh2020personalized}
& 89.63$\pm$4.25 & 86.89$\pm$2.81 & 72.28$\pm$1.96
& 59.83$\pm$0.62 & 54.05$\pm$3.58 & -- \\
\midrule

pFedKDH-Fair
& 91.06$\pm$0.63 & 85.81$\pm$0.15 & 74.32$\pm$0.33
& 67.39$\pm$0.27 & 60.22$\pm$0.38 & 41.39$\pm$0.50 \\

pFedKDH-EA
& \textbf{92.33$\pm$0.14} & \textbf{88.37$\pm$0.15} & \textbf{76.43$\pm$0.15}
& \textbf{68.45$\pm$0.47} & 60.51$\pm$0.30 & 41.96$\pm$0.85 \\
\bottomrule
\end{tabular}%
}

\end{table}

Table~\ref{tab:accuracy_mnist_fmnist} reports the final accuracy comparison on MNIST and Fashion-MNIST, while Table~\ref{tab:accuracy_cifar} reports the corresponding results on CIFAR10 and CIFAR100. Together, these tables summarize the final performance under different levels of statistical heterogeneity. pFedKDH obtains the highest mean accuracy in the most heterogeneous settings. For $\alpha=0.05$, it achieves the best mean accuracy on all four datasets. For $\alpha=0.10$, it remains the best method on MNIST, Fashion-MNIST, and CIFAR10, and is only 0.35 \% below FedALA on CIFAR100, while presenting a much lower standard deviation. The low standard deviations indicate that these gains are consistent across repetitions.

The base version, pFedKDH-Fair, also shows strong behavior without evaluation-time head adaptation. It is the second-best method in all MNIST settings, in Fashion-MNIST for $\alpha=0.05$ and $\alpha=0.10$, and in CIFAR10 for $\alpha=0.05$. Its standard deviation is also consistently small when compared with several personalized baselines. This confirms that the core mechanism of pFedKDH, namely backbone-only aggregation with persistent client-specific heads, already provides a stable personalization strategy.

The gains differ across baseline families. Compared with global baselines, pFedKDH improves the best FedAvg/FedProx result by 6.05 \% on MNIST, 24.47 on Fashion-MNIST, 37.66 on CIFAR10, and 32.71 on CIFAR100 at $\alpha=0.05$. Compared with regularization-based personalized methods such as Ditto and pFedMe, the gain reaches 8.62 \% on CIFAR100 with $\alpha=0.05$. Against backbone-head separation methods such as FedPer, FedRep, and FedBABU, pFedKDH still improves the best competing result by up to 1.82 \% on CIFAR10 with $\alpha=0.05$. The comparison indicates that head separation helps, while distillation adds further gains.

The cases where pFedKDH is not the best are concentrated in Fashion-MNIST with $\alpha=0.50$ and CIFAR100 with $\alpha=0.10$ and $\alpha=0.50$. In the first two cases, the mean gap is small and pFedKDH keeps lower variability. The larger gap on CIFAR100 with $\alpha=0.50$ suggests that many-class settings may require more class-aware teacher calibration or more adaptive knowledge transfer, since a single recalibrated teacher may become less informative when the label space is large and client label support is fragmented. The pFedMe result for CIFAR100 with \(\alpha=0.50\) is omitted because the run did not complete under the same computational budget.

All experiments use a fixed auxiliary set of $|D_s|=5000$ samples.
The sensitivity to $|D_s|$, class
imbalance within the auxiliary set, and domain shift between $D_s$
and the client distributions remains to be investigated. These factors
may affect the quality of teacher recalibration, particularly in
many-class and highly heterogeneous settings.

\vspace{-0.25cm}

\subsection{Component-wise Diagnostic Analysis}
\label{subsec:component_diagnostic_analysis}

We use a diagnostic ablation on CIFAR10 with \(\alpha=0.05\) to inspect the contribution of the main pFedKDH components. This experiment uses a single Monte Carlo repetition and is intended only as an explanatory analysis.

\begin{figure}[!h]
\centering
\includegraphics[
    width=0.4658\textwidth,
    height=0.4\textheight,
    keepaspectratio
]{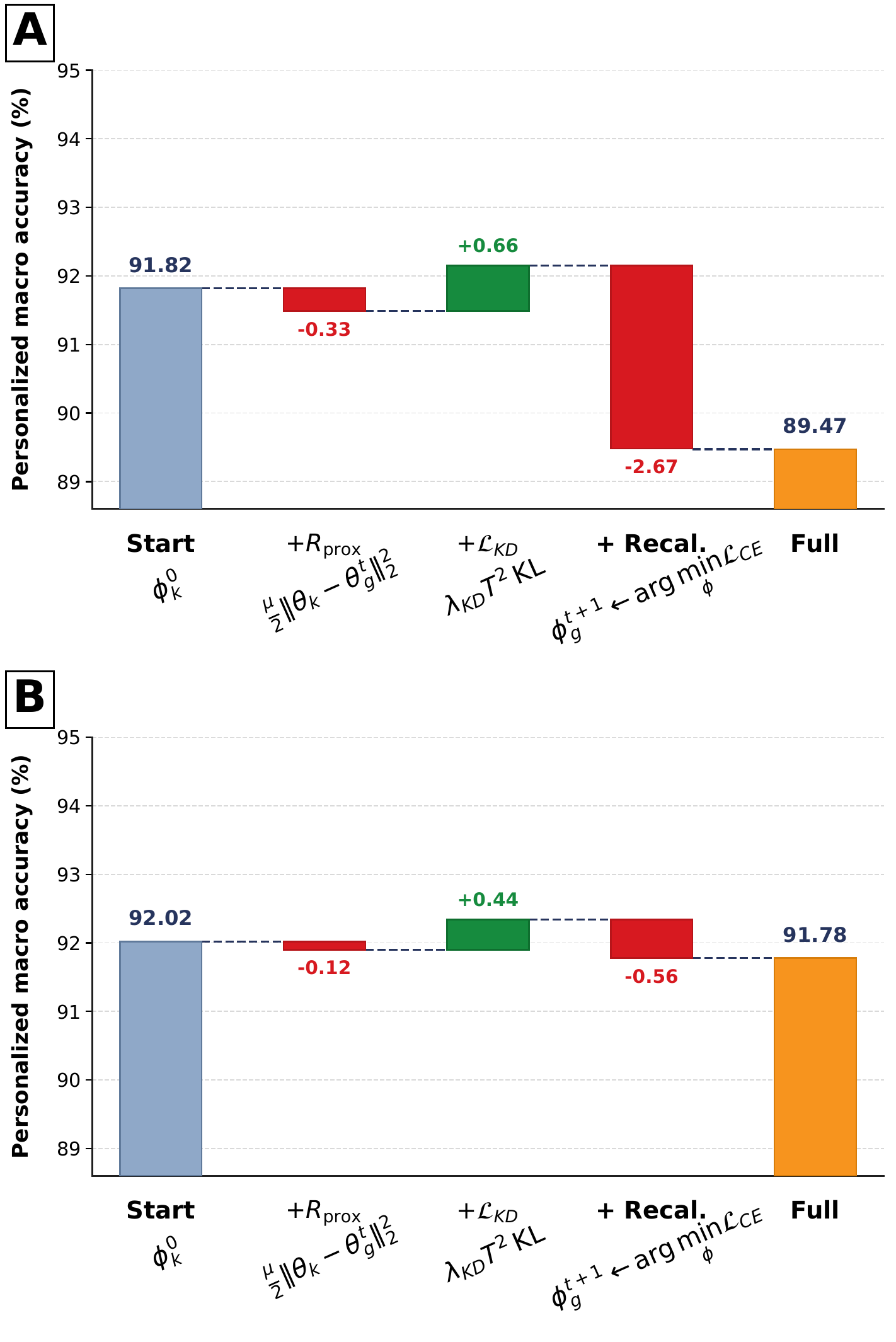}

\caption{Component-wise diagnostic ablation of pFedKDH on CIFAR10 under $\alpha=0.05$. Panel A shows pFedKDH-Fair, while Panel B shows pFedKDH-EA.}
\label{fig:ablation_pfedkdh}
\end{figure}

Figure~\ref{fig:ablation_pfedkdh}
shows that persistent local heads provide the main personalization gain, while knowledge distillation further improves local optimization. The proximal term has a small effect in this run. Recurrent recalibration mainly acts as a teacher-alignment mechanism, and its quantitative impact may vary with the dataset, class imbalance, and degree of label skew.

\subsection{Computational Cost}

Let $P_b$ and $P_h$ denote the numbers of backbone and head
parameters, respectively, and let $|S_t|$ be the number of clients
selected at round $t$. As only the backbone is uploaded, the
client-to-server communication cost per round is
$\mathcal{O}(|S_t|P_b)$, instead of
$\mathcal{O}(|S_t|(P_b+P_h))$ for full-model aggregation.
The server aggregation in (4) also requires
$\mathcal{O}(|S_t|P_b)$ operations. Teacher recalibration introduces
server-side optimization over $D_s$, but updates only the head while
the backbone remains fixed. Thus, pFedKDH preserves the same
backbone-order communication complexity as backbone-sharing PFL
methods, with additional computation arising mainly from local
distillation and teacher-head recalibration.

pFedKDH introduces a moderate computational overhead. pFedKDH-Fair and pFedKDH-EA require 6.029 s and 6.466 s per round, corresponding to 1.57\(\times\) and 1.68\(\times\) the runtime of the fastest method, FedPer. This cost is comparable to FedALA and FedBABU, and substantially lower than FedRep and pFedMe. 

\vspace{-0.25cm}

\section{Conclusions}
\label{sec:conclusion}

Statistical heterogeneity limits federated learning because a single shared classifier may not capture client-specific label distributions. pFedKDH addresses this by aggregating only the shared backbone, keeping persistent client-specific heads, and using a recalibrated global head as a teacher during local training.

Across the evaluated settings, pFedKDH achieves the best mean accuracy in most cases and remains competitive otherwise, while consistently showing low variability. Convergence and ablation results indicate that persistent heads preserve client-specific decision boundaries, while distillation improves global knowledge transfer under label skew.

Future work will investigate adaptive distillation, class-aware teacher
calibration, sensitivity to the size and class distribution of $D_s$,
robustness to domain shift, and protocols that reduce or eliminate
dependence on auxiliary server-side data.
\bibliographystyle{IEEEbib}
\bibliography{references}

@inproceedings{mcmahan2017communication,
  author    = {H. Brendan McMahan and Eider Moore and Daniel Ramage and Seth Hampson and Blaise Aguera y Arcas},
  title     = {Communication-efficient learning of deep networks from decentralized data},
  booktitle = {Proc. AISTATS},
  pages     = {1273--1282},
  year      = {2017}
}

@article{kairouz2021advances,
  author  = {Peter Kairouz and others},
  title   = {Advances and open problems in federated learning},
  journal = {Foundations and Trends in Machine Learning},
  volume  = {14},
  number  = {1--2},
  pages   = {1--210},
  year    = {2021}
}

@article{lu2024federated,
  author  = {Z. Lu and H. Pan and Y. Dai and X. Si and Y. Zhang},
  title   = {Federated learning with non-{IID} data: A survey},
  journal = {IEEE Internet of Things Journal},
  volume  = {11},
  number  = {11},
  pages   = {19188--19209},
  year    = {2024}
}

@article{chung2026decentralized,
  author  = {W.-C. Chung and C.-A. Lo and Y.-H. Lin and Z.-H. Chen and C.-L. Hung},
  title   = {Decentralized federated learning with non-{IID} data: Challenges, trends, and future opportunities},
  journal = {ACM Computing Surveys},
  volume  = {58},
  number  = {8},
  pages   = {Article 192},
  year    = {2026}
}

@article{ye2023heterogeneous,
  author  = {M. Ye and X. Fang and B. Du and P. C. Yuen and D. Tao},
  title   = {Heterogeneous federated learning: State-of-the-art and research challenges},
  journal = {ACM Computing Surveys},
  volume  = {56},
  number  = {3},
  pages   = {Article 79},
  year    = {2023}
}

@article{tan2022towards,
  author  = {A. Z. Tan and H. Yu and L. Cui and Q. Yang},
  title   = {Towards personalized federated learning},
  journal = {IEEE Transactions on Neural Networks and Learning Systems},
  year    = {2022}
}

@article{sabah2023model,
  author  = {F. Sabah and Y. Chen and Z. Yang and A. Raheem and M. Azam and R. Sarwar},
  title   = {Model optimization techniques in personalized federated learning: A survey},
  journal = {Expert Systems with Applications},
  year    = {2023}
}

@inproceedings{li2020federated,
  author    = {T. Li and A. K. Sahu and M. Zaheer and M. Sanjabi and A. Talwalkar and V. Smith},
  title     = {Federated optimization in heterogeneous networks},
  booktitle = {Proc. MLSys},
  year      = {2020}
}

@inproceedings{dinh2020personalized,
  author    = {C. T. Dinh and N. H. Tran and T. D. Nguyen},
  title     = {Personalized federated learning with {Moreau} envelopes},
  booktitle = {Advances in Neural Information Processing Systems},
  pages     = {21394--21405},
  year      = {2020}
}

@inproceedings{li2021ditto,
  author    = {T. Li and S. Hu and A. Beirami and V. Smith},
  title     = {Ditto: Fair and robust federated learning through personalization},
  booktitle = {Proc. ICML},
  pages     = {6357--6368},
  year      = {2021}
}

@article{arivazhagan2019federated,
  author  = {M. G. Arivazhagan and V. Aggarwal and A. K. Singh and S. Choudhary},
  title   = {Federated learning with personalization layers},
  journal = {arXiv preprint arXiv:1912.00818},
  year    = {2019}
}

@inproceedings{collins2021exploiting,
  author    = {L. Collins and H. Hassani and A. Mokhtari and S. Shakkottai},
  title     = {Exploiting shared representations for personalized federated learning},
  booktitle = {Proc. ICML},
  pages     = {2089--2099},
  year      = {2021}
}

@inproceedings{oh2022fedbabu,
  author    = {J. Oh and S. Kim and S.-Y. Yun},
  title     = {{FedBABU}: Toward enhanced representation for federated image classification},
  booktitle = {Proc. ICLR},
  year      = {2022}
}

@inproceedings{chen2022fedrod,
  author    = {H.-Y. Chen and W.-L. Chao},
  title     = {On bridging generic and personalized federated learning for image classification},
  booktitle = {Proc. International Conference on Learning Representations (ICLR)},
  year      = {2022}
}

@inproceedings{xiao2024fedloge,
  author    = {Z. Xiao and Z. Chen and L. Liu and Y. Feng and J. Wu and W. Liu and J. T. Zhou and H. H. Yang and Z. Liu},
  title     = {{FedLoGe}: Joint local and generic federated learning under long-tailed data},
  booktitle = {Proc. International Conference on Learning Representations (ICLR)},
  year      = {2024}
}

@inproceedings{tan2022fedproto,
  author    = {Y. Tan and G. Long and L. Liu and T. Zhou and Q. Lu and J. Jiang and C. Zhang},
  title     = {{FedProto}: Federated prototype learning across heterogeneous clients},
  booktitle = {Proc. AAAI},
  pages     = {8432--8440},
  year      = {2022}
}

@article{tan2023pfedsim,
  author  = {J. Tan and Y. Zhou and G. Liu and J. H. Wang and S. Yu},
  title   = {{pFedSim}: Similarity-aware model aggregation towards personalized federated learning},
  journal = {arXiv preprint arXiv:2305.15706},
  year    = {2023}
}

@inproceedings{zhang2023fedala,
  author    = {J. Zhang and Y. Hua and H. Wang and T. Song and Z. Xue and R. Ma and H. Guan},
  title     = {{FedALA}: Adaptive local aggregation for personalized federated learning},
  booktitle = {Proc. AAAI},
  pages     = {11237--11244},
  year      = {2023}
}

@inproceedings{yang2023fedfed,
  author    = {Z. Yang and Y. Zhang and Y. Zheng and X. Tian and H. Peng and T. Liu and B. Han},
  title     = {{FedFed}: Feature distillation against data heterogeneity in federated learning},
  booktitle = {Advances in Neural Information Processing Systems},
  volume    = {36},
  pages     = {60397--60428},
  year      = {2023}
}

@article{hinton2015distilling,
  author  = {G. Hinton and O. Vinyals and J. Dean},
  title   = {Distilling the knowledge in a neural network},
  journal = {arXiv preprint arXiv:1503.02531},
  year    = {2015}
}

\end{document}